\documentclass[a4paper]{article}
\usepackage[utf8]{inputenc}
\usepackage[T1]{fontenc}
\usepackage{lmodern}
\usepackage{xcolor}
\definecolor{fgreen}{rgb}{0.1,0.5,0.2}
\definecolor{cgray}{rgb}{0.2,0.6,0.2}
\definecolor{marine}{rgb}{0,0.2,0.6}
\usepackage[hidelinks]{hyperref}
\usepackage{graphicx}
\graphicspath{{./}{./figs/}}
\newcommand{\figScale}{0.425}
\newcommand{\twographics}[2]
  {\includegraphics[trim = 20 0 20 20, clip, scale = \figScale]{#1}
   \includegraphics[trim = 20 0 20 20, clip, scale = \figScale]{#2}}
\newcommand{\twoscplots}[2]
  {\includegraphics[trim = 2 0 2 2, clip, scale = 0.21]{#1}
   \includegraphics[trim = 2 2 2 0, clip, scale = 0.18]{#2}}
\usepackage{listings}
\lstnewenvironment{lstcode}[1][]{\lstset{language=C++,#1}}{}
\newcommand{\lsti}{\lstinline}

\usepackage{filecontents}
\newcommand{\mytitle}{ML Supported Predictions for SAT Solvers Performance}
\hypersetup{
  pdfinfo={
    Title={\mytitle},
    Author={A.-M. Leventi-Peetz, J.-V. Peetz, M. Rohde}
    }
}

\newcommand{\inst}[1]{{$^#1$}}
\newcommand{\email}[1]{\texttt{#1}}
\newcommand{\keywords}[1]{\\[2ex]{\bf Keywords:} #1}
\begin{document}
%

\title{\mytitle}

\author{A.-M. Leventi-Peetz\inst{1} \and Jörg-Volker Peetz \and Martina Rohde\inst{1}}

\date{\small $^1$ Federal Office for Information Security,\\ Godesberger Allee 185--189, DE-53175 Bonn, Germany\\
  \email{leventi-peetz@bsi.bund.de}
}

\maketitle

\begin{abstract}
  In order to classify the indeterministic termination behavior of the open source SAT solver CryptoMiniSat in multi-threading mode while processing hard to solve boolean satisfiability problem instances, internal solver runtime parameters have been collected and analyzed. A subset of these parameters has been selected and employed as features vector to successfully create a machine learning model for the binary classification of the solver's termination behavior with any single new solving run of a not yet solved instance. The model can be used for the early estimation of a solving attempt as belonging or not belonging to the class of candidates with good chances for a fast termination. In this context a combination of active profiles of runtime characteristics appear to mirror the influence of the solver's momentary heuristics on the immediate quality of the solver's resolution process.  Because runtime parameters of already the first two solving iterations are enough to forecast termination of the attempt with good success scores, the results of the present work deliver a promising basis which can be further developed in order to enrich CryptoMiniSat or generally any modern SAT solver with AI abilities.
\end{abstract}

\section{Introduction}
The significance of SAT\footnote{SAT -- \emph{satisfiablity}} solvers as a core technology for the analysis, synthesis, verification and testing of security properties of hardware and
software products is established and well known.  Automated Reasoning
techniques for finding bugs and flaws examine nowadays billion lines
of computer code with the use of Boolean and Constraint Solvers on domains
of interest.  However due to the continuous improvement of SAT solver
efficiency during the last decades and the dramatic scalability of
these solvers against large real-world formulas, many new
application cases arise in which SAT solvers get deployed for tackling
hard problems, which were believed to be in general intractable but
yet get solved~\cite{ganesh2018MLSat}.  SAT solvers are also known to
be used for security protocol analysis
\cite{vanhoef2018SymbExec,armando2006SATModelChecking} as well as
for tasks like the automated verification of access control policies,
automatic \emph{Anomaly Detection} in network configuration policies
\cite{lorenz2015PolAnomaly} and verification of general Access Control
Systems where access rules are first encoded in SAT representation
\cite{hughes2008AutoVerifACP,mauro2017anomaly}.  Furthermore
SAT-based cryptanalysis methods are in advance and report increasing
successes, like for example cryptographic key recovery by solving
instances which encode diverse cipher attacks
\cite{kehui2011CMSAlgebraicSCAttack,semenov2018CryptoAttSAT,otpuschenn2016CryptoSAT,lafitte2014SATCryptanalysis}
etc.

Also the solution of constraint optimization problems for real-world
applications on the basis of already very competitive MaxSAT solvers,
the great majority of which are core-guided, heavily relying on the
power of SAT solvers, is the best way of proving unsatisfiability of
subsets of soft constraints, or unsat cores, in an iterative way
towards an optimal solution \cite{jarvisalo2016SatAI}.  New
algorithmic solutions instantiating innovative approaches to solve
various data analysis problems in ML,\footnote{ML -- \emph{machine learning}} like correlation
clustering, causal discovery, inference etc. are based on SAT and
Boolean optimization solvers.  Decision and optimization problems in
artificial intelligence profit by the application of SAT solvers
\cite{rintanen2013AIplaSAT} but also the opposite direction is
pursued, namely the improvement of SAT solving using ML
\cite{devlin2008satclass,liang2016LRB,liang2018MLSAT}.  The intrinsic connection of the two
subjects, SAT solvers and AI, seems to be growing especially also
under the scope of global trends intending to incorporate AI and
ML technologies in the majority of the next generation cybersecurity
solutions. The improvement of SAT solver performance by means of AI is
a topic of intense and general interest and motivated this work.

\section{Organization of this paper and contributions}
First, a brief account is given here of previously gained experience with
CryptoMiniSat \cite{Soos2009CMS}, gathered while studying the solver's behavior when
engaged for the solution of hard CNF instances representing KPA in cryptanalysis.\footnote{CNF -- \emph{conjunctive normal form}; KPA -- \emph{known-plaintext attacks}} Similar instances whose
solution is far from being trivial, are used also for the results
produced in this work \cite{leventi2018CMSparopt}. However these instances are here taken as an
example of especially hard instances to solve while the achieved
results should be relevant to the solution of arbitrary hard instances
having similar features to those of the here employed ones.  In what
follows the motivation of the present work is substantiated and then
the procedure followed to produce our results as well as the results
themselves will be presented.  In the last part we summarize about
this work and discuss about further investigations planned.  In the
past the CMS\footnote{CMS -- \emph{CryptoMiniSat}} solver's performance has been studied by carrying out
runtime tests both with the solver in default configuration and with
various solver switches set.  The tests showed that the solver runtime
until the solution is found, in case the job doesn't previously stop
because of some time-limit setting, is subjected to distinct
statistical variations. This is due to the indeterministic behavior of
the solver in multi-thread operation mode. The complexity of the here
discussed problems though excludes one-thread operation from being an
option. The performed runtime tests were in average highly time
consuming both when a termination was reached and when the job had to
be interrupted because it reached some previously defined upper run
time limit. The job interruption practice was motivated by an amplitude
of experience showing that if some long runtime limit has been
surpassed without a solution found, the majority of test runs do not
terminate at all. As a matter of fact, even under identical solver
parameter configuration when running several tests to solve one and
the same instance, one cannot avoid diverging solving times or finding
no solution at all for instances which are solvable by construction.
All cryptanalytic instances used in experiments possess by
construction one single solution, which was in this case the sought-for
cryptographic key.  The performance of a large number of tests had
allowed a statistical analysis of the non-deterministic solver
runtimes to empirically define command line parameter combinations for
CMS which yield best runtimes medians for the type of instances taken
under examination.  The application of an AAC\footnote{AAC -- \emph{automatic algorithm configuration}} tool which followed \cite{SMACv3-2017}, allowed a
systematic exploration of the configuration parameter space reaching
beyond the empirical tests to discover even better configuration
parameter combinations.  The results of those efforts have been quite
encouraging, demonstrating a lowering of the median of runtimes by
30\,\% to finally 90\,\% with the application of the AAC \cite{leventi2018CMSparopt}. The limitation we see however in further pursuing this
approach is that when a semi-automatic tuning of the solver's
configuration parameters is to be performed, this has to be carried
out on the basis of previously cumulated values of best achieved
runtimes. Those best runtimes are of course obtainable at a high
computational and time cost which should probably have to be
repeatedly afforded, in case the expensively discovered effective
configuration parameter settings are not globally valid but rather
problem specific.

\subsection{Contributions of the present work}
In the present work we show how to create a prognosis concerning the
successful termination of a solving process not by properly adjusting some solver
configuration parameters but by using internal runtime parameters of
the beginning of the process during the process. Instead of analyzing
effectiveness of the solver's configuration in a post-hoc manner, that
is following the event of numerous successful terminations, we have
here chosen to observe and analyze the joint evolution of solver
internal parameters that are dynamically changing during the automatic
state transformation of the instance while the solver is searching for
a solution. These parameters which are issued by the solver when
running in verbose mode, have been at first separately investigated in
order to see if they demonstrate any correlation to the duration of
job-runtimes and final successful termination. This question could not
get uniquely answered.  This circumstance motivated the tryout of a
subset of the solver's dynamically changing runtime parameters as
model features for building an ML model with the purpose to
detect if the solver's active state evolution follows a direction with good
chances to terminate timely or not.  Taking into consideration the
fact that the joint observation of parameters that belong to the two
or three first iterations of the solving process showed to be sufficient
to construct the ML model so as to get some decent classification
results, this approach can be seen as a workable basis to later devise
and incorporate an internal mechanism in the solver to trigger early
changes of search strategy on the basis of internal short-termed \emph{collective} parameter
changes at a minimal time cost. Given the fact that the time
length of the solver's iterations normally essentially grows with the
iteration's number order, the limitation of the needed for the model
runtime parameters on those of the first iterations, definitely helps
avoid waste of resources on hopeless solving efforts involving many
iterations. The innovation of the effort described here lies also in
the fact that a combination of parameter profiles originating from
both terminating and non-terminating test runs equally contributes to
the model building. The ultimate task is building an ML model to be
used for an almost real-time quality estimation of the solver's
momentary search direction.

To our knowledge the here presented way of joint employment of
solver's internal statistics in order to create solver-forecasts has not been
attempted before.

\section{Solver runtime parameters: description, analysis, and illustration}
We have selected runtime parameters which we consider to be in general
significant for characterizing the state of the solving process, and
this not only in relation to the here regarded CNF instances.
The parameter charts plotted in the graphics below are
representative for the processing of instances of similar features.
As features of the here employed CNF instances we observe the number of variables $L$, the number of CNF clauses $N$, the length of clauses (minimum, lower quartile, median, upper quartile, and maximum length), the sum of all occurrences of all variables, the fraction of clauses by length in the CNF instance, the occurrence of literals in clauses, and their mean value \cite{hutter2014SATfeat}.
The three instances whose runtime parameters are plotted here belong to three different variations of the same mathematical problem (known-plaintext attack on the round-reduced AES-64 model cipher) with instance densities $N/L$ assuming the values 303.4 (18-vs), 304.2 (20-vs) and 306.6 (30-vs).
We could prove that similar parameter graphics are produced for instances created on the basis of the same mathematical problem but with different problem parameters (different number of plaintexts, for our instances 18, 20, and 30, respectively and/or different key).

The parameter names are almost self-explanatory in the context of the
functionality of a CDCL\footnote{CDCL -- \emph{conflict driven clause learning}} solver.  We have experimented with two
slightly different sets of parameters for the creation of the ML model
and these sets are displayed in the first and second columns of table~\ref{tab:params},
where the abbreviation \lsti|props| stands for propagations.
\begin{table}[htbp]
  \caption{\label{tab:params}Runtime parameters as model features}
  \vspace{1ex}
  \begin{minipage}{\columnwidth}
  \centering
  \begin{tabular}{l|l}
    \hline
    Set 1 & Set 2 \\
    \hline
    \lsti|all-threads| & \lsti|all-threads| \\
    \lsti|conflicts/second| & \lsti|conflicts/second|\\
    \lsti|blocked-restarts| & \lsti|blocked-restarts| \\
    \lsti|restarts|         & \lsti|restarts| \\
    \lsti|props/decision|   & \lsti|props/decision| \\
    \lsti|props/conflict|   & \lsti|literals/conflict| \\
                            & \lsti|decisions/conflict| \\
    \hline
   \end{tabular}
  \end{minipage}
\end{table}

\begin{figure}[htbp]
  \centering
  \twographics{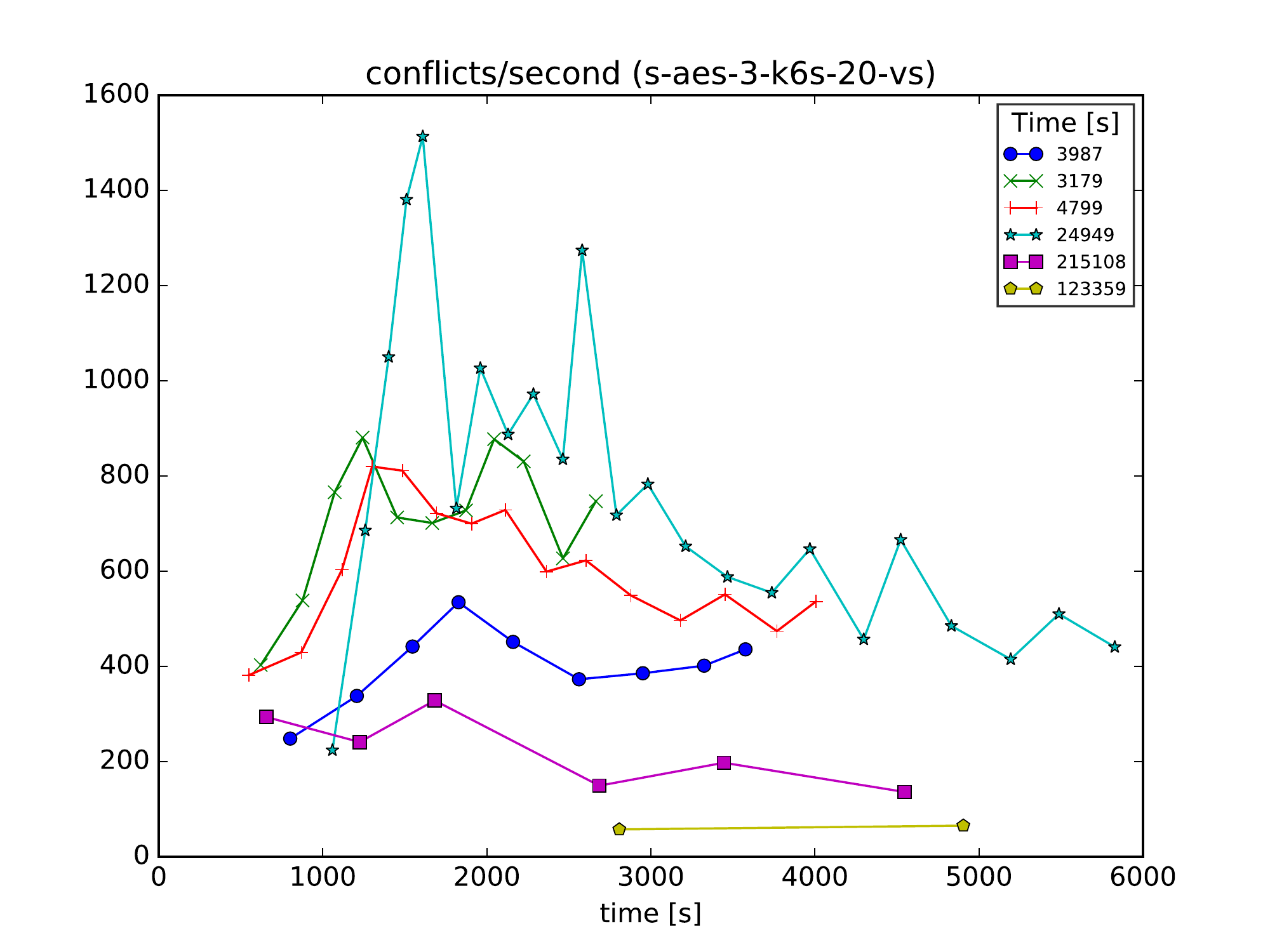}{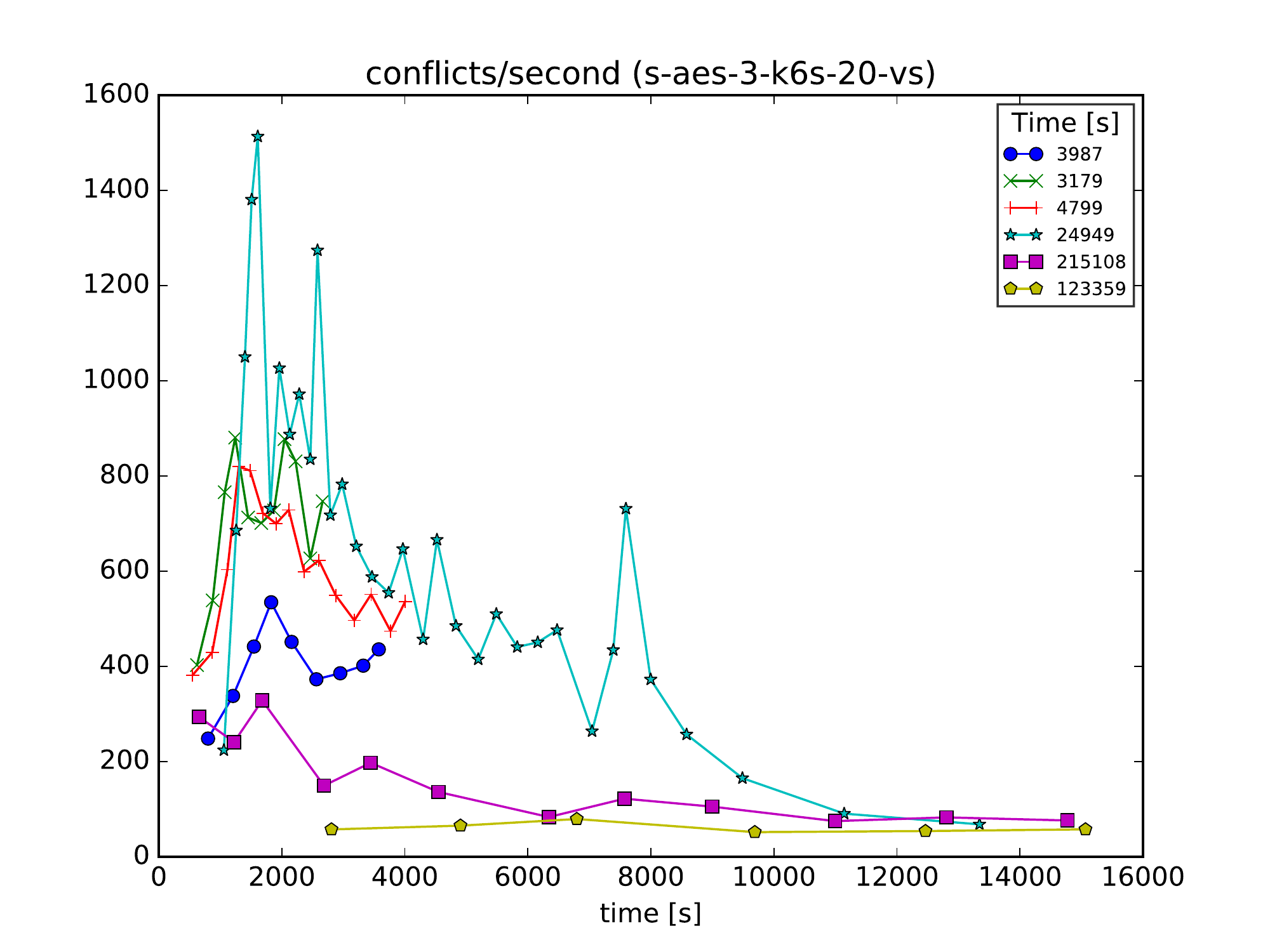}
  \caption[Fig. 1ab]
  {\label{f:Fig1ab} parameter evolutions for the instance 20-vs}
\end{figure}

The parameter \lsti|all-threads| expresses the total time consumed by a
solver iteration calculated by adding the solving times contributed
by the various threads during this iteration. All here discussed
parameters are regularly delivered by the solver at the end of each and every
iteration, with the solver running in verbose mode. In Figure~\ref{f:Fig1ab} the
\lsti|conflicts/second| parameters for six different runs of the same
instance that terminated at different times with the solver
identically configured for all runs, is exemplarily depicted. There
exist differences in the structure, especially the heights and in the
positions of the peaks of the curves which seem to associate to the
length of the runtime until a successful termination of the
corresponding job (see legends in Figure~\ref{f:Fig1ab}) has been reached.
%
%

It seems plausible to attribute to the steep rising number of
conflicts at the beginning of the solving-process (that means during
the first and/or the second iterations) an early generation of a lot
of additional useful informations for the solver.  This event
accelerates the learning effect which would justify the assumption of
a direct association of a shorter solution runtime to the occurrence
of many conflicts at the beginning of the solving process. Many
conflicts per second along the search path naturally results in a
shorter iteration time, as there is a limit of conflicts per
iteration, allowed by the solver, which in this case gets earlier
reached. Advantageous are therefore short iteration times, arising out
of the occurrence of high conflict rates during short iterations.

The markers on the curves correspond to the begin/end of an iteration.
A short first iteration time sets the begin of the corresponding plot
closer to the y-axis and it is notable that curves starting far to the
left describe evolutions of solutions with mostly shorter therefore
better runtimes.

Curves which despite their starting far to the left do evolve to
describe long and therefore bad termination runtimes, usually
demonstrate also early negative or very flat gradients as regards the
\lsti|conflicts/second| parameter.  To this, one can compare the two
relatively flat curves with termination times 215,108 and 123,359
seconds, respectively (see legend of Figure~\ref{f:Fig1ab}).  These two curves
could also be categorized as not successfully terminating runs, if an upper time-limit
had been set for the allowed runtime of the corresponding jobs.

\begin{figure}[htbp]
  \centering
  \twographics{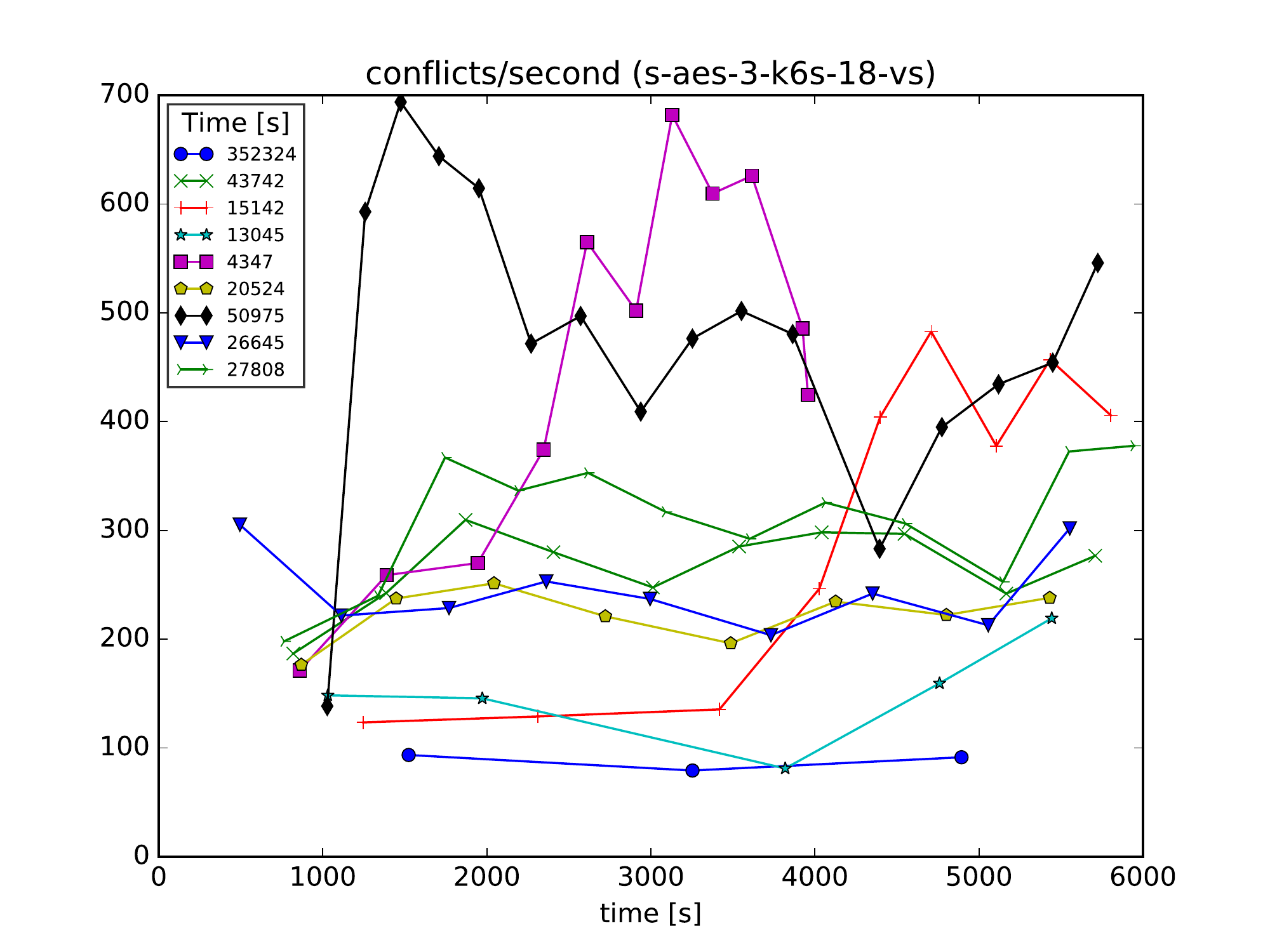}{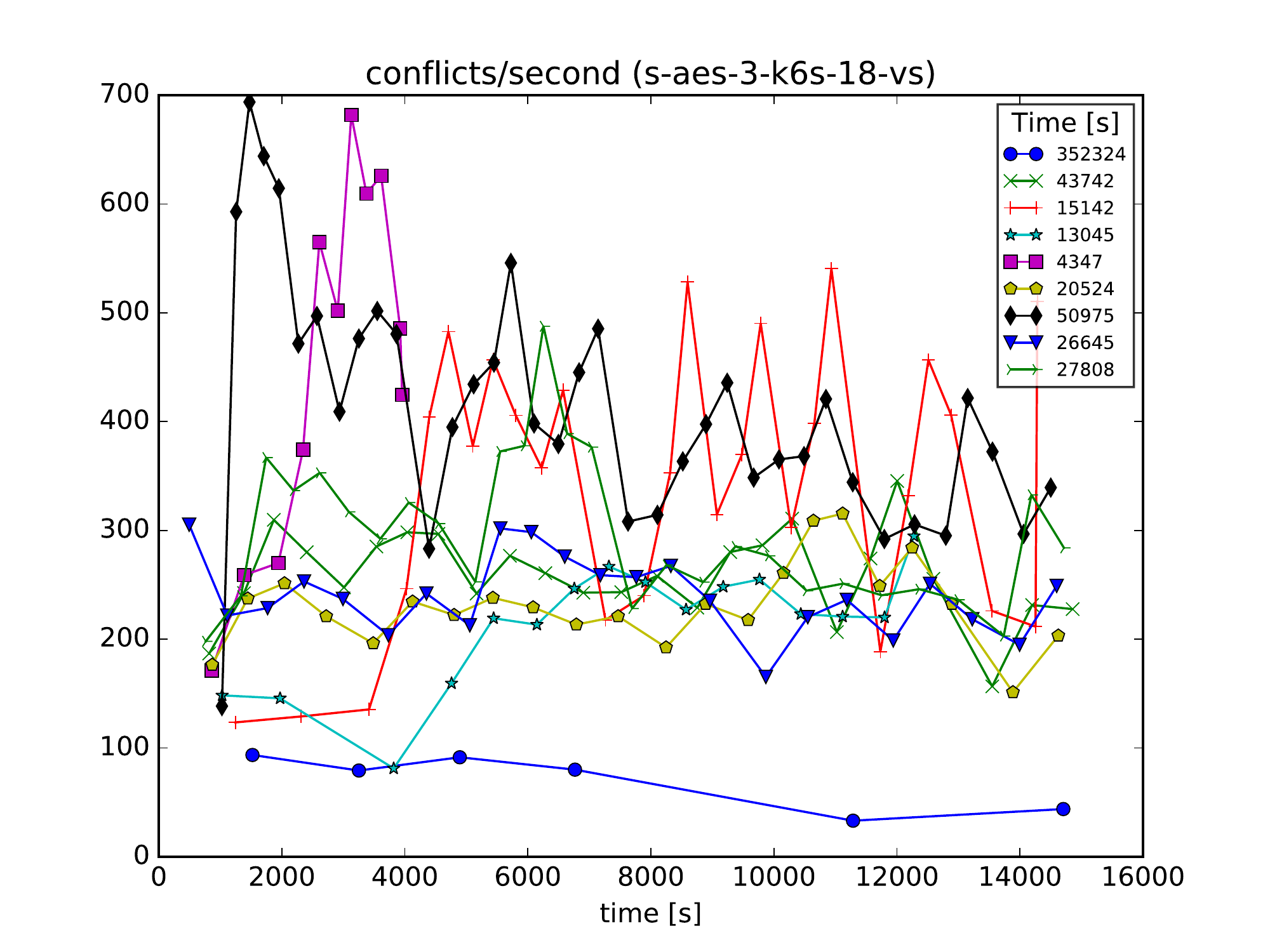}
  \caption[Fig. 2ab]
  {\label{f:Fig2ab}parameter evolutions for the instance 18-vs}
\end{figure}

\begin{figure}[htbp]
  \centering
  \twographics{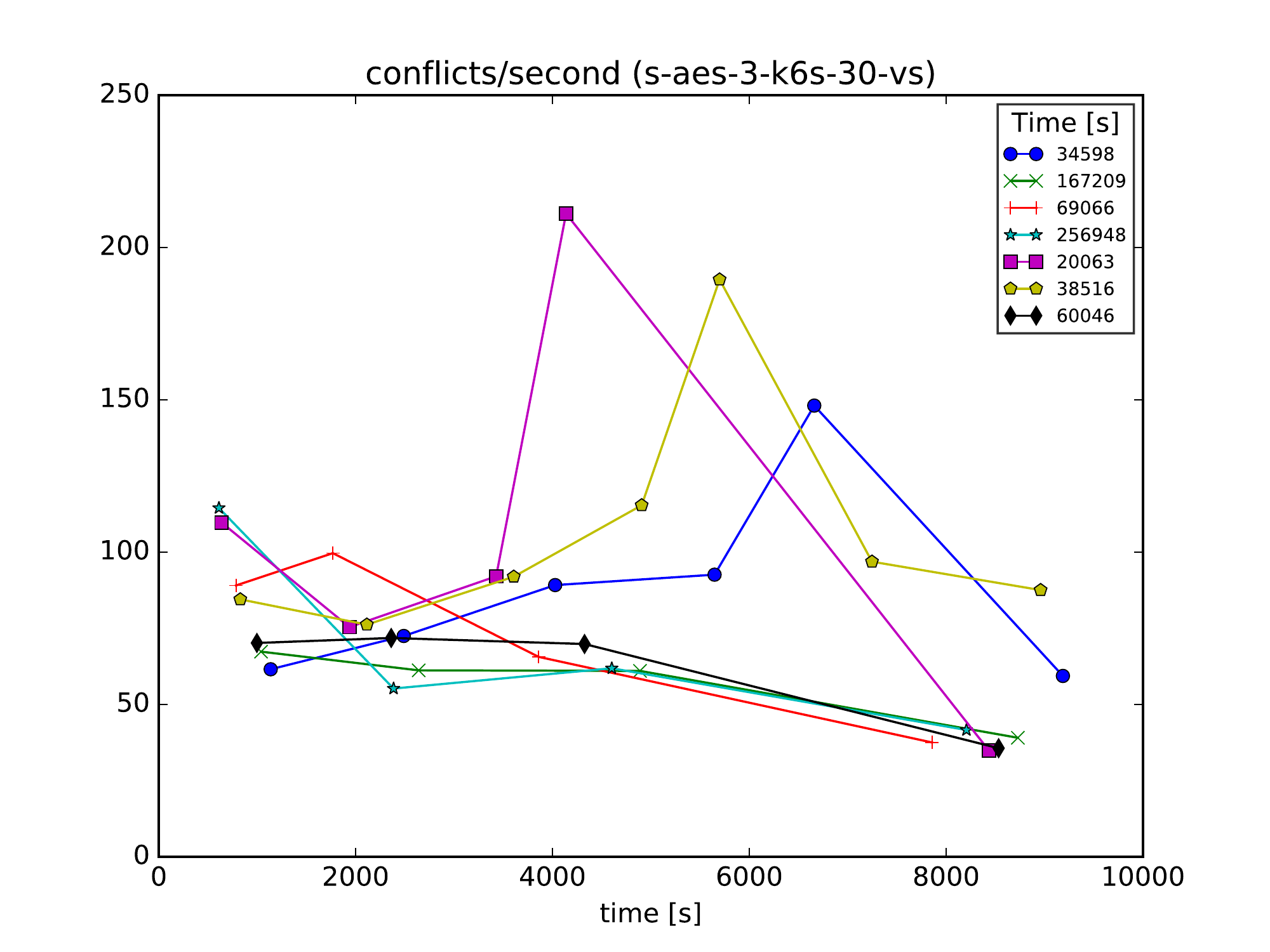}{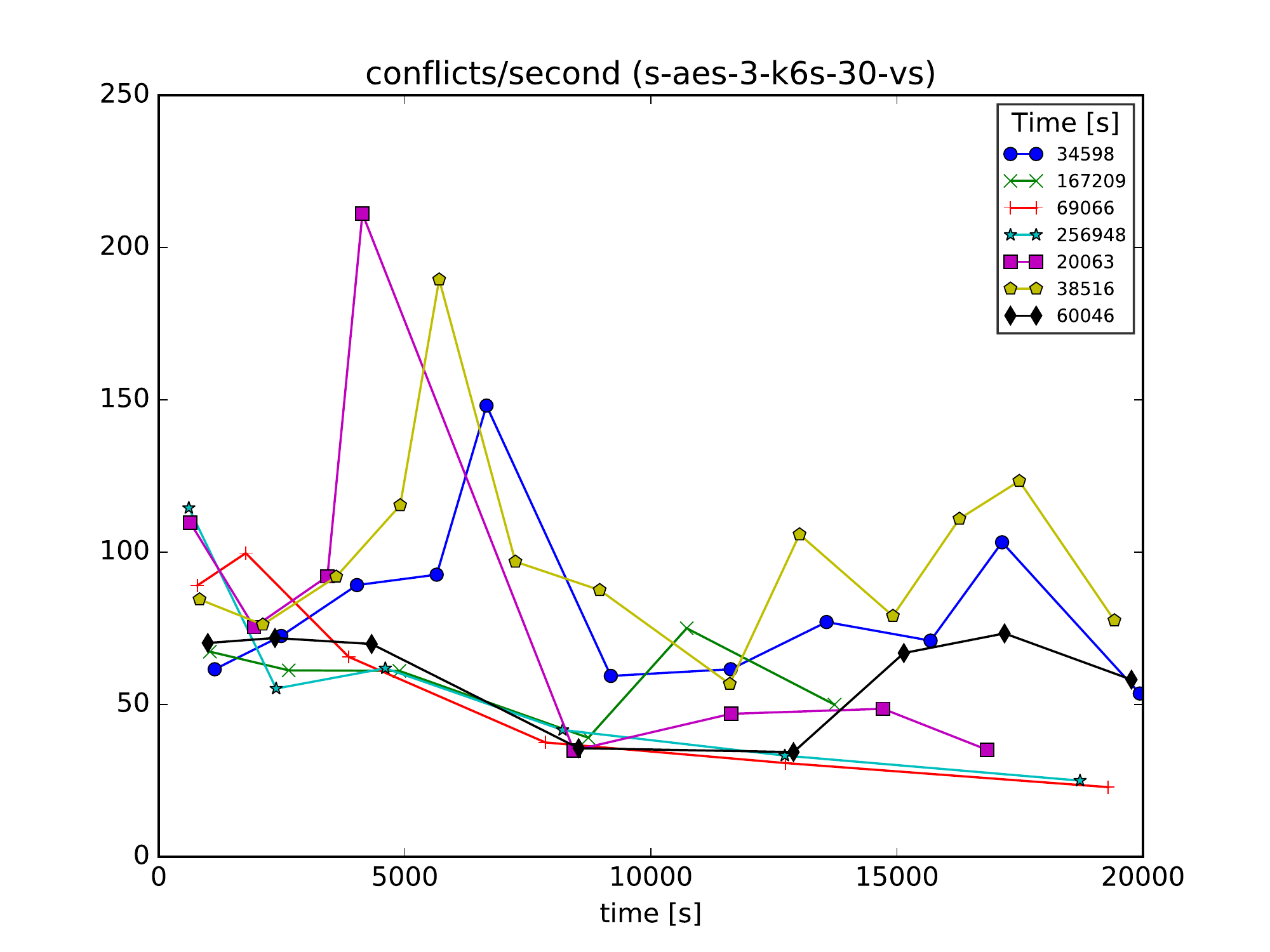}
  \caption[Fig. 3ab]
  {\label{f:Fig3ab}parameter evolutions for the instance 30-vs}
\end{figure}

Comparable results are depicted in Figure~\ref{f:Fig2ab}
(Figure~\ref{f:Fig3ab}) where the evolution of nine (seven) different runs of
the instance 18-vs (30-vs) are shown.  The instance in
Figure~\ref{f:Fig2ab} is the most difficult to solve in comparison to
the other two instances of Figure~\ref{f:Fig1ab} and
Figure~\ref{f:Fig3ab} respectively. In this more difficult case one observes
several intensive or less intensive learning phases in a sequence,
represented by more than one distinct peaks in the corresponding
curve plots. Fine differences in the start values seem to play also a
considerable role for the consequent runtime evolution. Not only the
duration of the first iteration but also the initial value of the
parameter \lsti|conflicts/second| is in this respect
significant. Striking is again the appearance of the late starting and
thoroughly flat curve corresponding to the very late termination time of
352,324 seconds (see legends of Figure~\ref{f:Fig2ab}).

%
%
%

%
%

Considering the fact that no continuous correlation
connecting any of the runtime parameters with the corresponding solver
termination time is to be found, the question arises if one can use
statistical traits of the joint evolution of parameters in a
combination, not in order to discover conditions for some optimal
solution time but in order to distinguish between a solver path leading to a
termination and one which does not.

The second runtime parameter here checked is that of the \lsti|restarts|.
Restarts is a critically important heuristic according to many
experts in the field of CDCL SAT-solver research.  Restart schemes
have been evaluated in detail and a particular benefit could be
identified in frequent restarts when combined with \emph{phase
  saving}~\cite{biere2015cdclrestarts}. Restarts are considered to \emph{compact the assignment
  stack} and frequent restarts enhance the quality of the learnt
clauses thus shorten the solution time.  In a recent work an
ML-based restart policy has been introduced to trigger a restart every
time an unfavorable forecast regarding the quality of the expected
new to be created learned clauses arises\cite{liang2018MLSAT}.

\begin{figure}[htbp]
  \centering
  \twographics{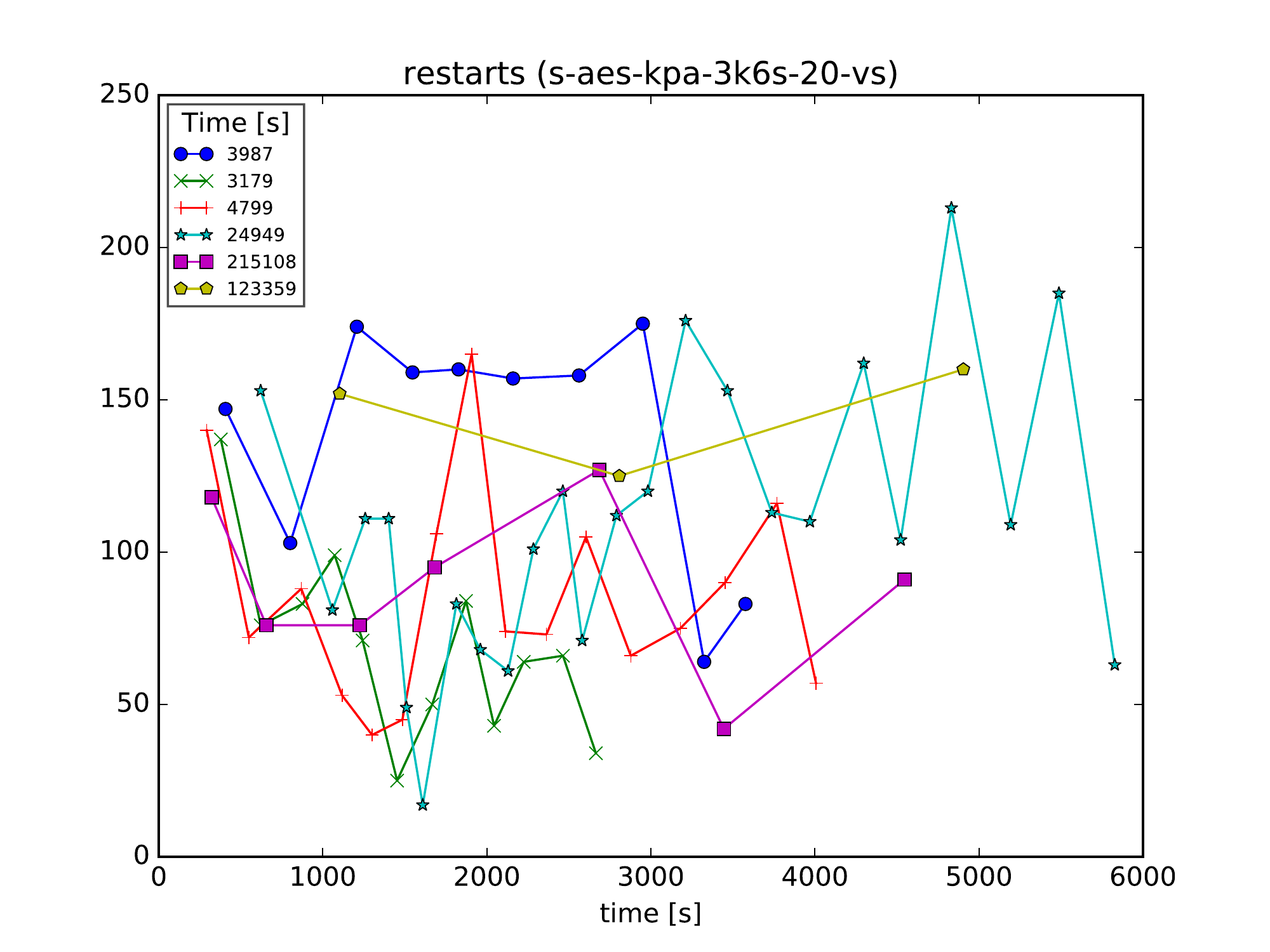}{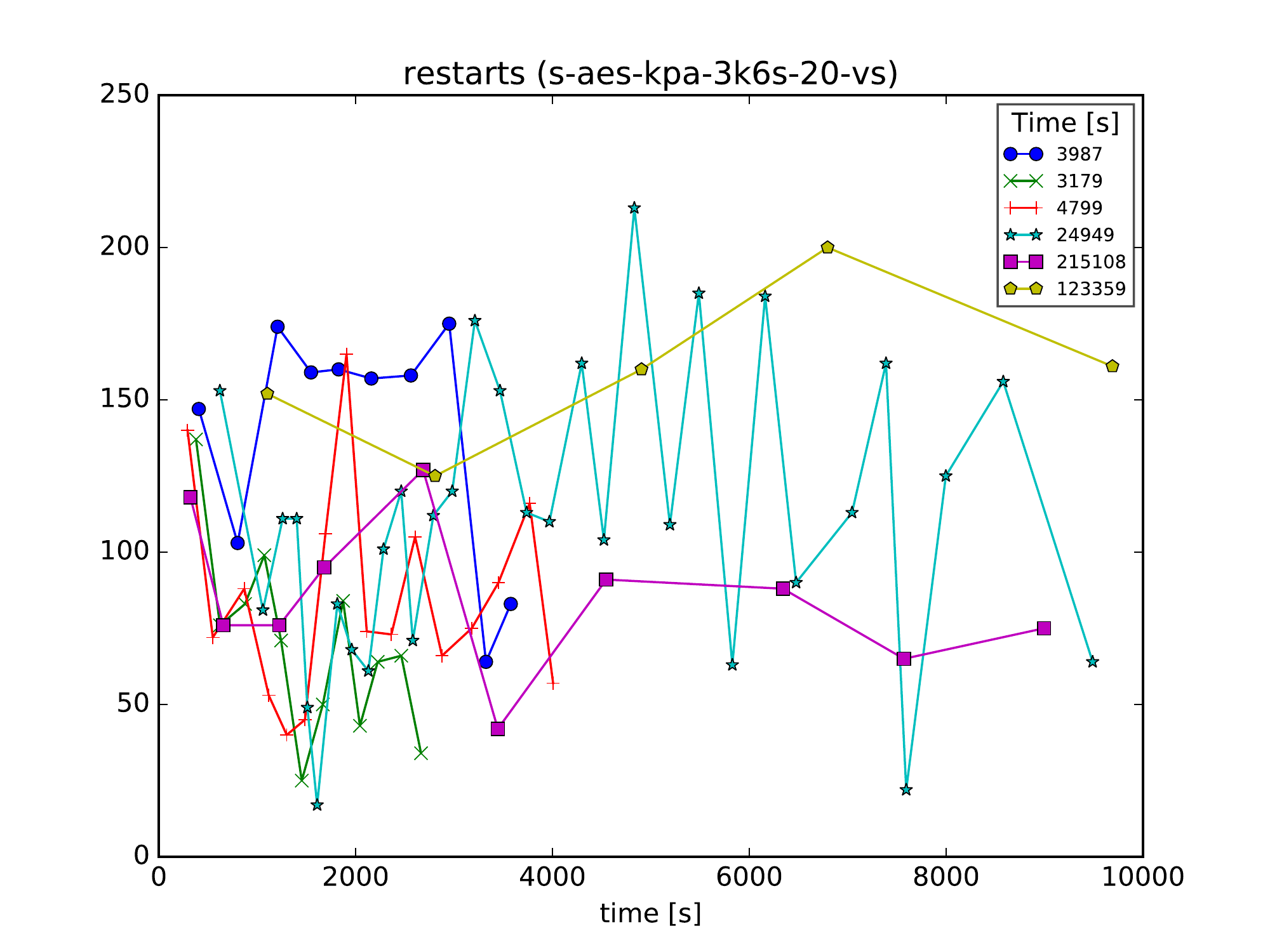}
  \caption[Fig. 4ab]
  {\label{f:Fig4ab}restarts for the instance 20-vs}
\end{figure}

\begin{figure}[htbp]
  \centering
  \twographics{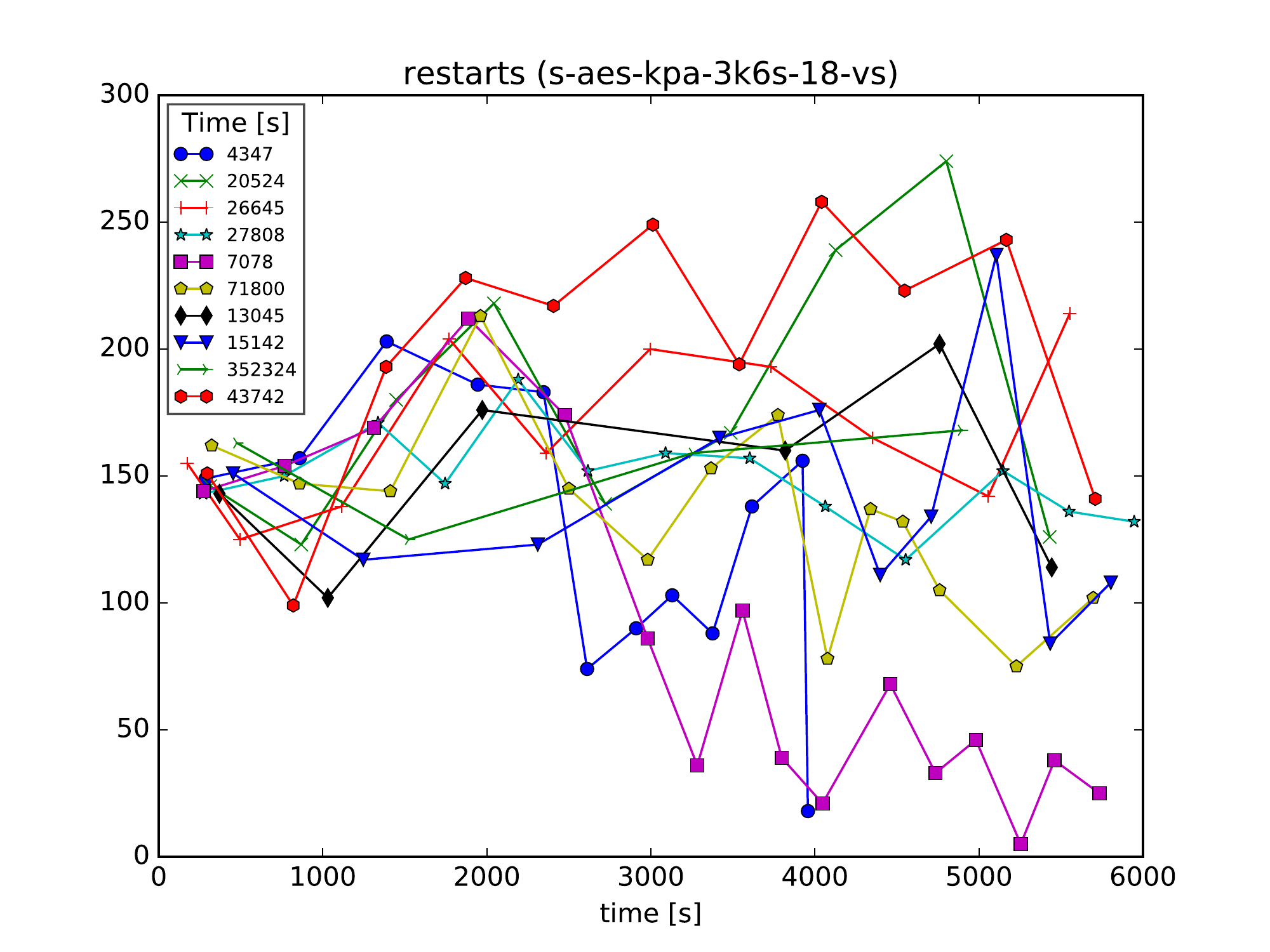}{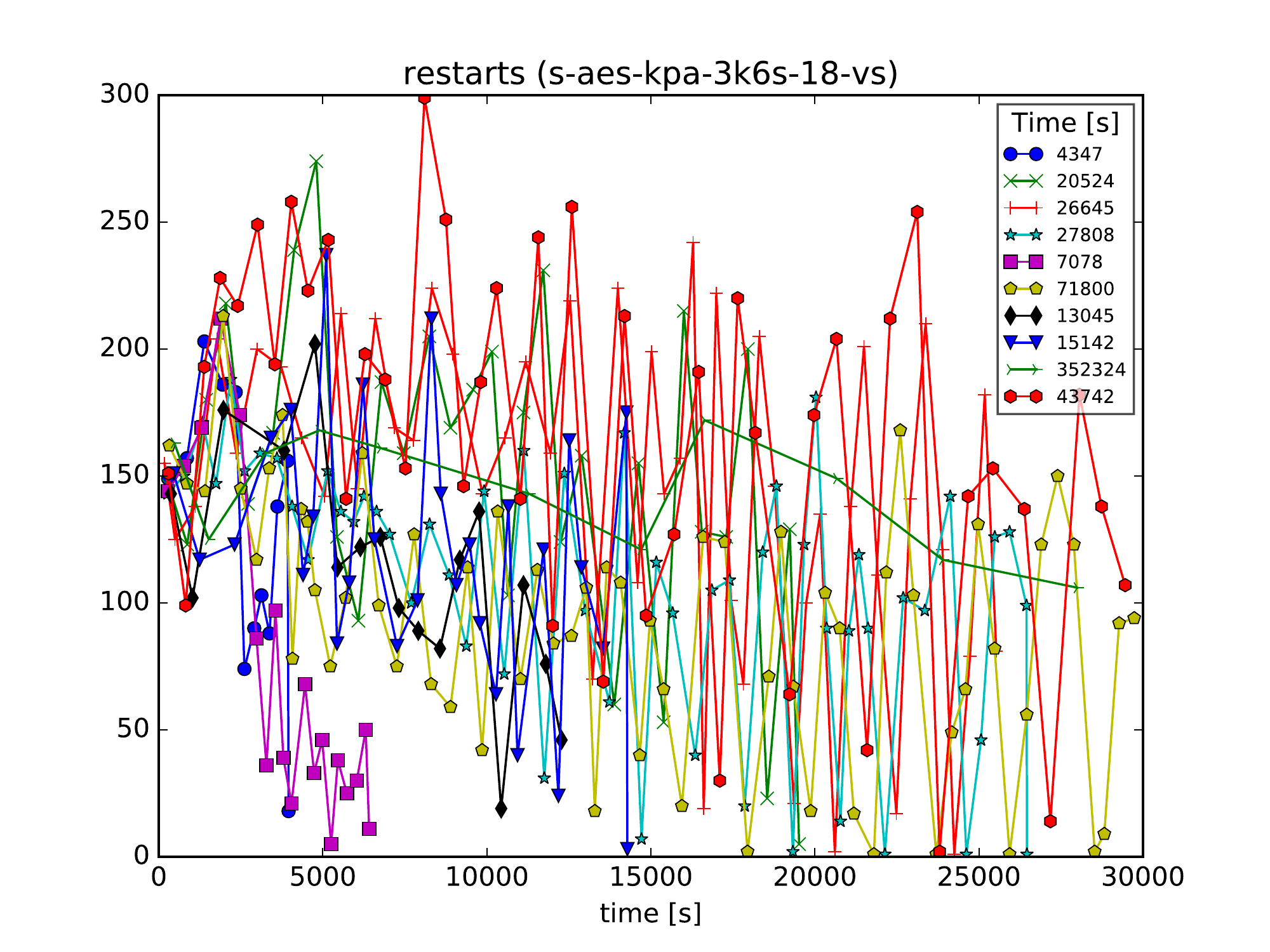}
  \caption[Fig. 5ab]
  {\label{f:Fig5ab}restarts for the instance 18-vs}
\end{figure}

\begin{figure}[htbp]
  \centering
  \twographics{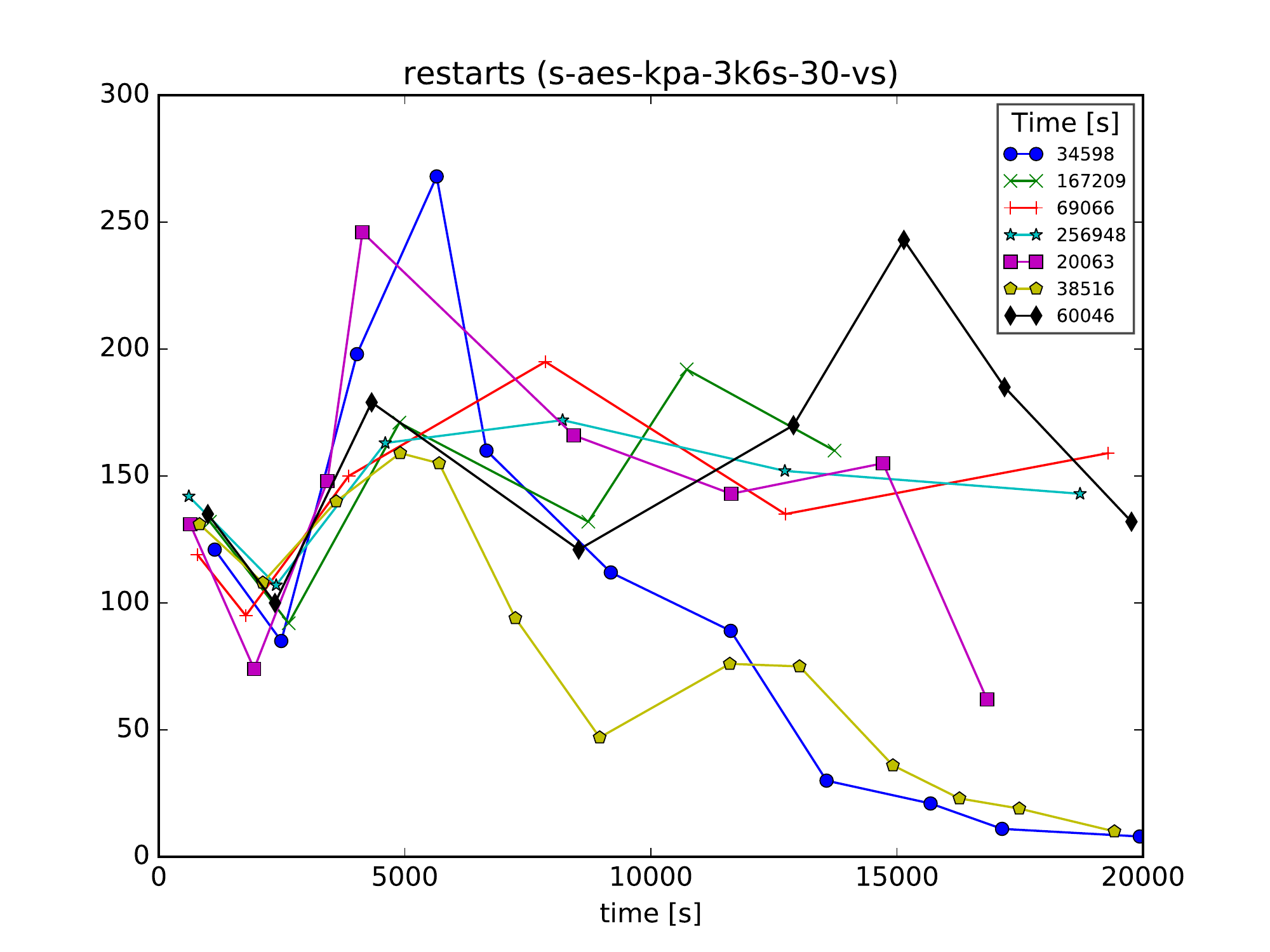}{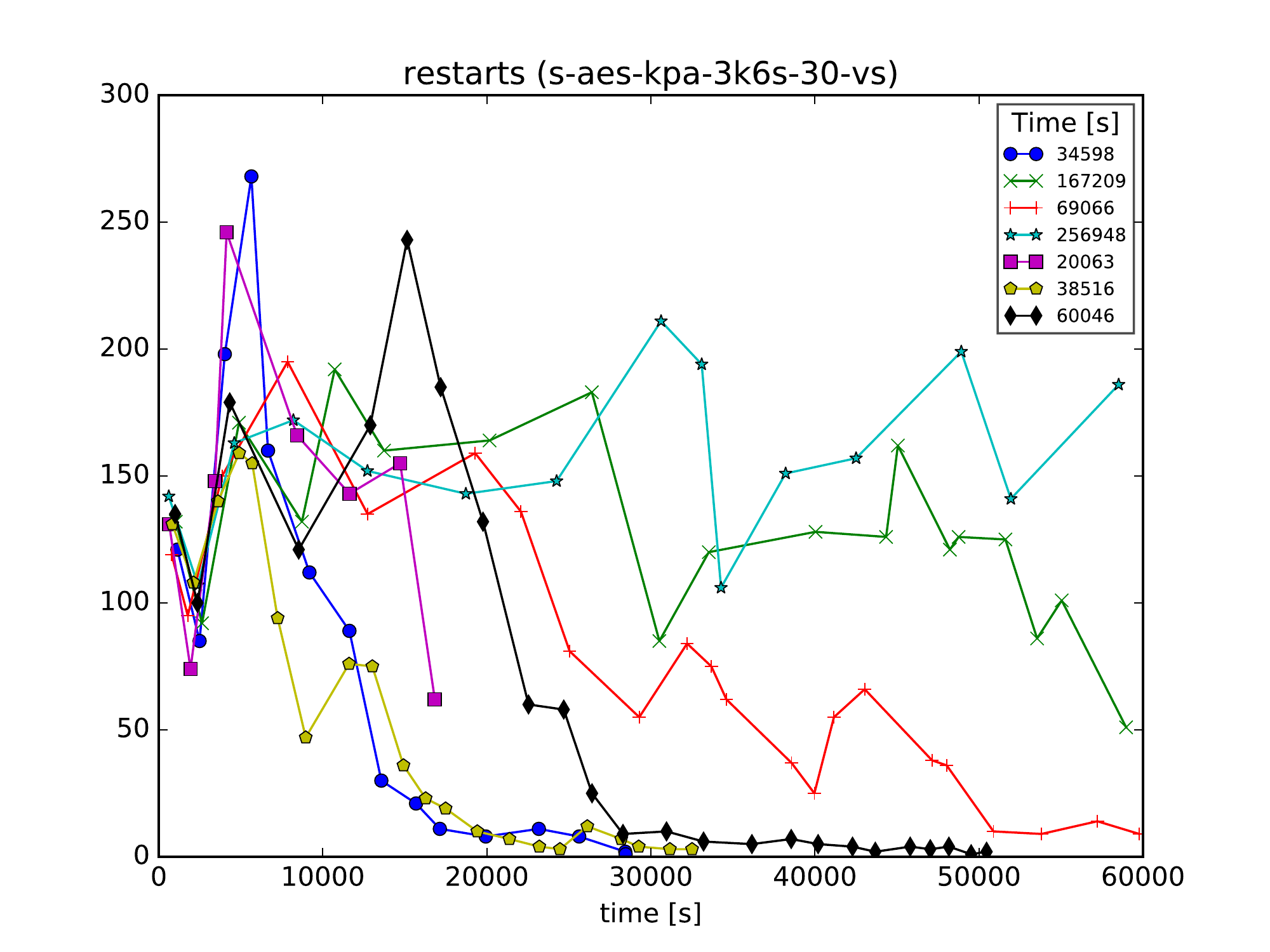}
  \caption[Fig. 6ab]
  {\label{f:Fig6ab}restarts for the instance 30-vs}
\end{figure}

In Figures~\ref{f:Fig4ab}, \ref{f:Fig5ab} and \ref{f:Fig6ab} there are
depicted the courses of the corresponding \lsti|restarts| parameters for
the same runs whose \lsti|conflicts/second| parameters have been
plotted above.

%
%
%
%
%
%

A comparison between these new three plots reveals as a common
indicator of unpromising (not timely expected to terminate) runs, the
corresponding poorly structured, and at parts continuously evolving
course of a curve.  This \emph{adverse} curve-shape feature, if
observed alone, becomes especially obvious only when watched over a
time period which is longer than two or three iterations.  In
combination with the iteration length, given by the \lsti|all-threads|
parameter though, the \lsti|restarts| parameter can indeed count
as a criterium to forecast a timely termination during
the first two iterations alone.

Similar remarks apply well for \lsti|blocked-restarts|, the complementary parameter, and the rest of the parameters taken for the ML model building, whose plots are not displayed here out of space considerations.
In conclusion, one can ascertain that a missing \emph{agility} in the timely evolution of runtime parameters and especially in combination with a low conflict rate at the beginning of the solving process, plausibly signalize little chance for a timely termination of a solving process.

\begin{figure}[htbp]
  \centering
  \twoscplots{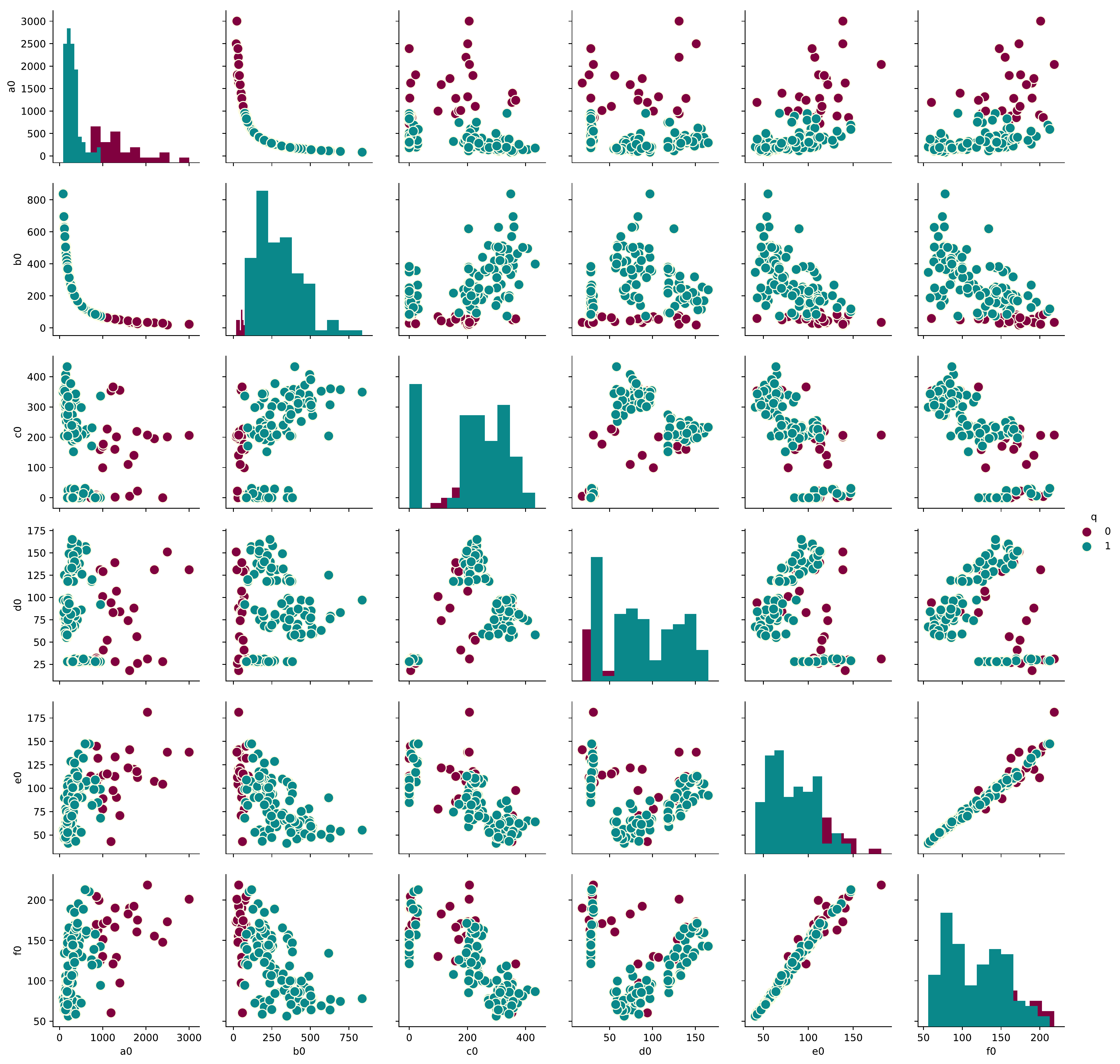}{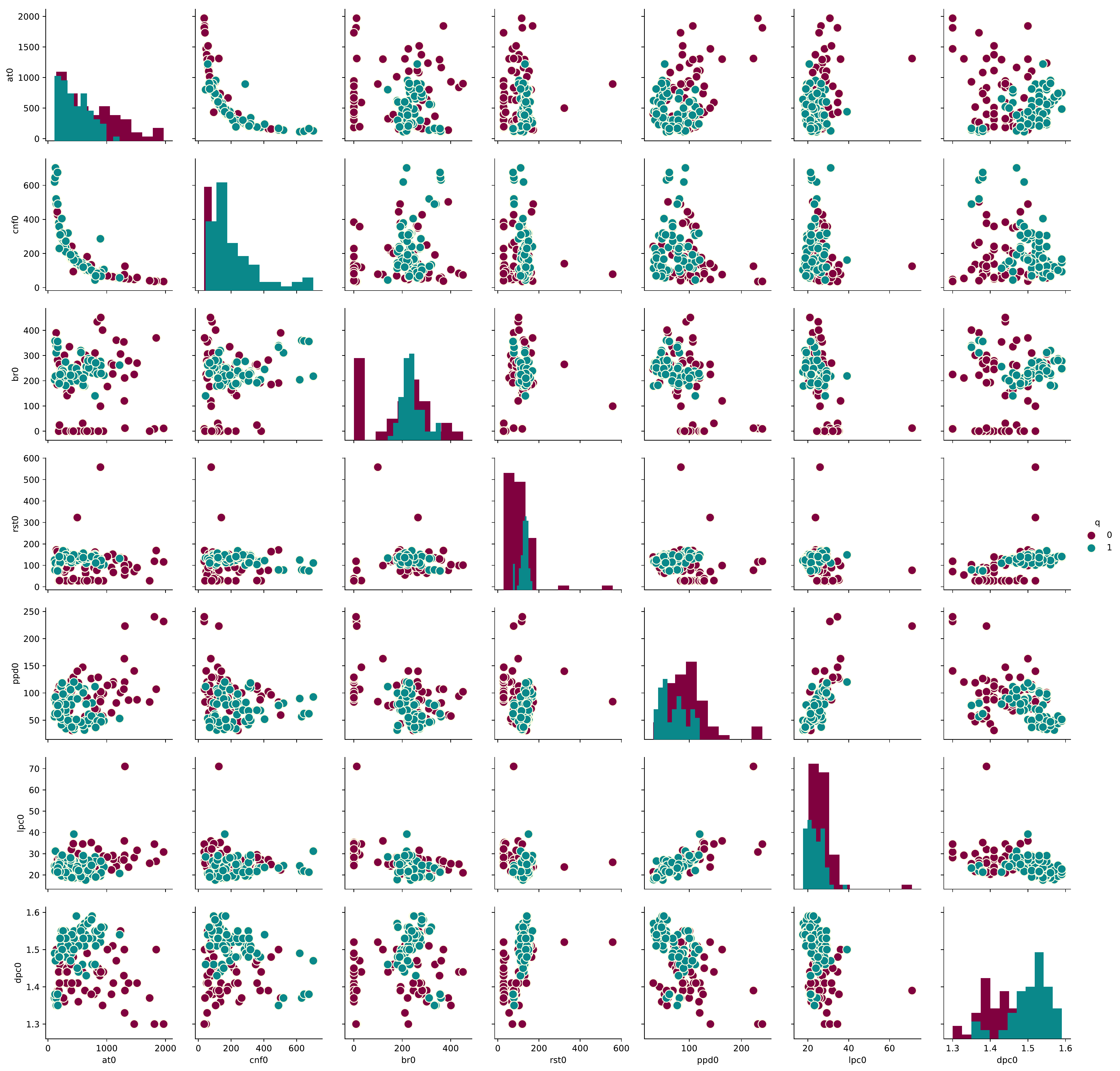}
  \caption[Scatterplots]
  {\label{f:scatplot}Pairwise scatterplots of six (top) and of seven (bottom) CMS runtime parameters. Each point indicates a CMS run. Colors indicate the classification of the test run.}
\end{figure}

Figure~\ref{f:scatplot} shows pairwise scatterplots of the six (top) and seven (bottom) runtime parameters (compare first and second columns respectively of table~\ref{tab:params}. Also different CNF instances were used for each plot). These parameters correspond to the first of the two iterations taken for building the ML model.
The plots help to reveal pairwise correlations between the model features. They indicate also the intricacy implied in finding an analytical function for the classification. Points with different colors symbolize different classification tags.

%


\section{ML models with CMS-runtime parameters as model features}
In order to investigate the feasibility of correctly and fast predicting the finding of a solution with the CMS solver on the basis of an ML model, we used CMS-runtime parameters of previous test-run cases to construct  NN\footnote{NN -- \emph{neural network}} classifiers. For the practical implementation we have employed the high-level Keras framework~\cite{chollet2015keras}. Keras is an open source neural-network library written in Python. It is capable of running on top of TensorFlow, Cognitive Toolkit (CNTK), or Theano. It was developed with a focus on enabling fast experimentation.

The training datasets for the models were compiled by extraction of the runtime parameters out of log-data originating in a set of test run cases, incorporating two equal numbers of terminating and not terminating processes (150 all together for each CNF instance) and which were then accordingly tagged with 1 or 0.
We explored three different NN models for binary classification, each of them has been trained with the same training datasets. The test datasets had in each case half the magnitude of the corresponding training dataset (75 test cases).

The first NN model we tried was also the simplest one consisting of mainly two layers of neurons. For the first layer, the input layer, there was taken a number of neurons equal to the runtime parameters of the CMS, planned to serve as features for the model. See table~\ref{tab:params}.
One and the same runtime parameter was considered as a different model feature if belonged to a different iteration, so that we had 12 (14) input parameters for the model. The second layer (output layer) had just one neuron and delivered the binary classification. 

The second NN model had an intermediate layer added with half as much neurons as the first layer, here six. The additional, hidden layer helps an NN-model to better approximate the classification of non-linearly separable data.

The third NN model was derived from the second model by dropping a percentage of the input values for the second and the third layer which should help prevent overfitting. This model is called a multilayer perceptron for binary classification. In all NN-models non-linear activation functions were used.

Independent of model, when employing the 6 features of the first column of table~\ref{tab:params} we had for all three instance-cases an equally good
hit ratio near 90\,\% while the formal model accuracy was 100\,\%. With the seven parameters as model features of the second column of table~\ref{tab:params} these simple models deliver less success with a hit score of about 70\,\%.
This shows that parameter (feature) selection is very important. Also, a more developed NN model might be necessary for a reliable classification prediction.
These first results encourage the building of further more sophisticated models in the future.

\section{Conclusions and future work}
The SAT solver CryptoMiniSat reenacting produces a series of runtime parameter values while processing a CNF instance until solving or failing to solve it within a time limit. We showed that it is possible to select and employ subsets of these runtime parameters, the same for all runs, and use them as features to build an ML model for solver runtime classification. For any future attempt to solve a new CNF instance of features similar to those of the instance used to create the training data for the model, this model enables the early classification of the attempt as a fast or a late terminating run.  By early is meant that the parameter profiles released during the first two solver iterations are already sufficient for a classification of this solving attempt.

In a future work the model should be extended so as to become capable to generate forecasts for attempts to solve problems bigger than the ones which it has been trained for. Also the possibility to extend the model so as to generate forecasts for CNF instances of any features and especially for hard random instances of any parametrization lies in the focus of our plans.
\clearpage
\bibliographystyle{plain}
\bibliography{\jobname}
\end{document}